\documentclass[conference]{IEEEtran}
\IEEEoverridecommandlockouts

\usepackage{cite}
\usepackage{amsmath,amssymb,amsfonts}
\usepackage{algorithmic}
\usepackage{graphicx}
\usepackage{textcomp}
\usepackage{xcolor}
\usepackage{bbding}
\usepackage{booktabs}
\usepackage{balance}
\usepackage{subfigure}
\usepackage{color}
\usepackage{tabularx}
\usepackage{threeparttable}
\def\BibTeX{{\rm B\kern-.05em{\sc i\kern-.025em b}\kern-.08em
    T\kern-.1667em\lower.7ex\hbox{E}\kern-.125emX}}

\begin{document}

\title{{Improving Multi-Vehicle Perception Fusion with Millimeter-Wave Radar Assistance}
\thanks{\IEEEauthorrefmark{1}Wei Wang is the corresponding author.}
}

\author{
	\IEEEauthorblockN{Zhiqing Luo\IEEEauthorrefmark{2}, Yi Wang\IEEEauthorrefmark{2}, Yingying He\IEEEauthorrefmark{2}, and Wei Wang{\IEEEauthorrefmark{3}\IEEEauthorrefmark{1}}}
	
	\IEEEauthorblockA{\IEEEauthorrefmark{2}\textit{Huazhong University of Science and Technology}, \IEEEauthorrefmark{3}\textit{Wuhan University}\\
		Email: \{zhiqing\_luo, yi\_wang\_, heyingying, weiwangw\}@hust.edu.cn}}

\maketitle

\begin{abstract}
Cooperative perception enables vehicles to share sensor readings and has become a new paradigm to improve driving safety, where the key enabling technology for realizing this vision is to real-time and accurately align and fuse the perceptions. Recent advances to align the views rely on high-density LiDAR data or fine-grained image feature representations, which however fail to  meet the requirements of accuracy, real-time, and adaptability for autonomous driving. To this end, we present MMatch, a lightweight system that enables accurate and real-time perception fusion with mmWave radar point clouds. The key insight is that fine-grained spatial information provided by the radar present unique associations with all the vehicles even in two separate views. As a result, by capturing and understanding the unique local and global position of the targets in this association, we can quickly find out all the co-visible vehicles for view alignment. We implement MMatch on both the datasets collected from the CARLA platform and the real-world traffic with over 15,000 radar point cloud pairs. Experimental results show that MMatch achieves decimeter-level accuracy within 59ms, which significantly improves the reliability for autonomous driving.
\end{abstract}


\section{Introduction}\label{sec:introduction}
\begin{figure}[t]
	\centering
	\includegraphics[width=1\linewidth]{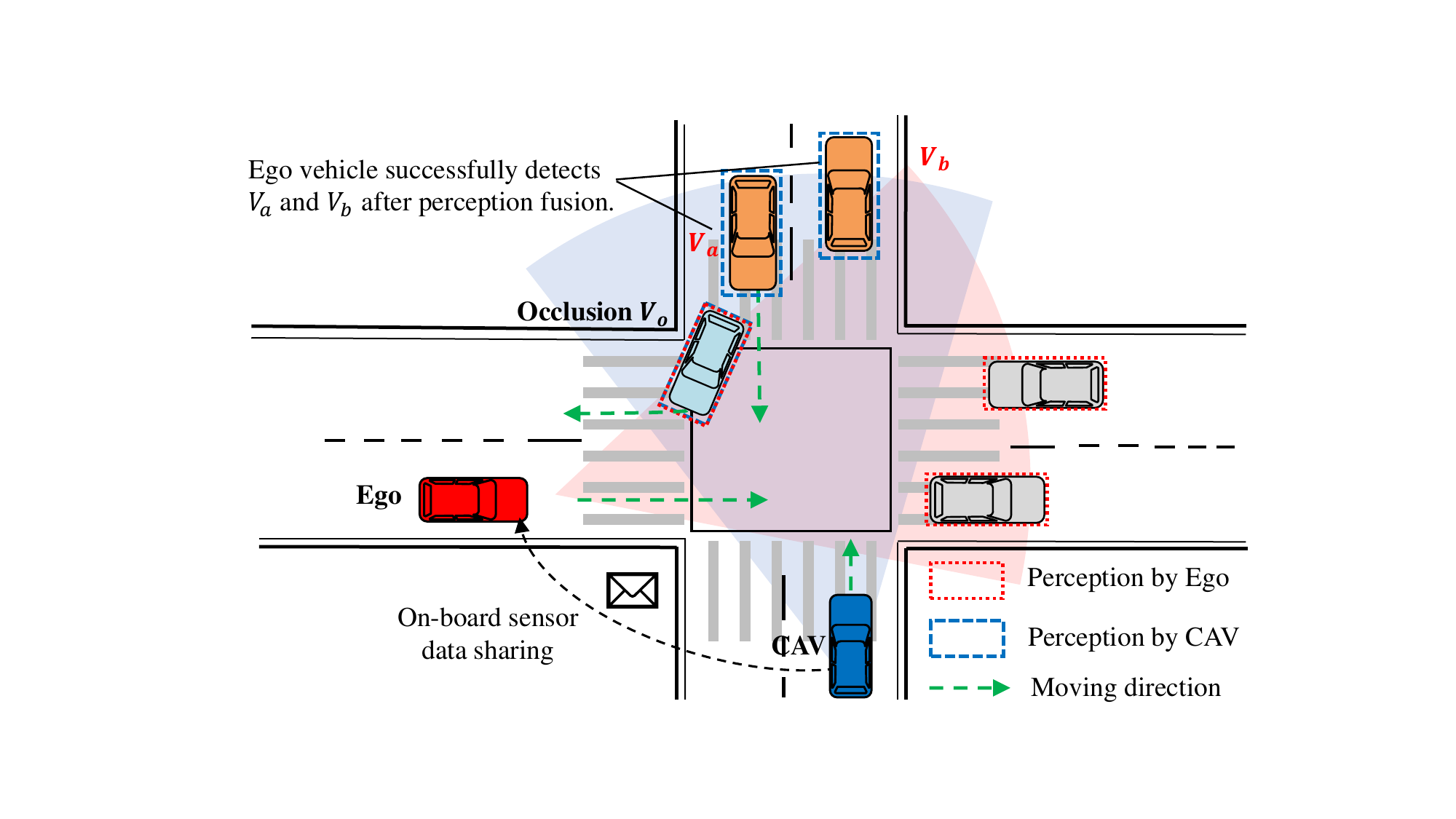} 
	\caption{An illustration of cooperative perception in autonomous driving. }
	\label{fig:illustration}\vspace{-0.2cm}
\end{figure}

Autonomous driving has been considered a promising technology in transforming our daily transportation. To ensure the reliability and safety, a variety of sensors like LiDAR, cameras, and radar have been deployed to sense and understand the surrounding traffic~\cite{bresson2017simultaneous}. However, single-vehicle perception systems suffer from barrier occlusion and limited perception range, which significantly compromise the understanding of the surrounding situation and safe planning~\cite{qiu2018avr}. To break this limitation, cooperative perception has enabled a new paradigm, where multiple connected and automated vehicles (CAVs) share and align the sensor data to improve the perceiving capability~\cite{caillot2022survey,xiao2023toward}. As an example shown in Fig.~\ref{fig:illustration}, fusing the sensor information shared from the cooperative CAV, the Ego vehicle successfully sees across the occlusion.

View alignment is the crux in the cooperative perception pipeline, where the Ego vehicle requires finding the transformation matrix to align the separate views. However, compared with the SLAM system in a single agent, designing a robust and effective alignment scheme is more challenging from separate vehicles, which needs to meet the requirements along three fronts~\cite{caillot2022survey}:
\begin{itemize}
	\item \textbf{Accuracy.} To satisfy the various autonomous driving tasks, such as path planning, lane changes, and overtaking, the alignment scheme should achieve decimeter-level registration accuracy.
	\item \textbf{Real-time.} Limited by communication bandwidth, time latency for data sharing and alignment requires to be short, where the processing needs to be completed within tens of milliseconds to avoid catastrophic accidents.
	\item \textbf{Adaptability.} Dynamic environments and viewing angle differences have significant impacts on the alignment performance, and the perception fusion scheme is expected to be well-adaptive in these challenging situations.
\end{itemize}

Unfortunately, no system exists today that can meet all these requirements simultaneously. Traditional approaches propose to align and merge the sensor data by relying on GNSS-based pose estimation~\cite{arnold2020cooperative}, which however achieve an average error of several meters, and are inapplicable for many driving tasks. To address this issue, recent study aims to share a small number of landmark keypoints of the ground signs in the images, which achieves high localization accuracy while meets the real-time requirement.  However, the textures on the ground lack spatial information and are easily occluded by vehicles, especially in crowded traffic. Thanks to the super-density nature, recent LiDAR-based alignment systems consider either sacrificing a lot of time to share the raw data for accurate merge~\cite{he2021vi} or extracting heterogeneous representations in the dense LiDAR point clouds (PCDs) for low latency~\cite{shi2022vips}. Nevertheless, denser LiDARs require a higher maintenance need and are susceptible to adverse conditions, such as dust and smog~\cite{raj2020survey}. Compared with these two light-based sensors, low-cost and all-weather enabled millimeter-Wave (mmWave) radars are more capable in challenging environments~\cite{gao2023robust}, and provide the crucial information of all targets’ velocities and positions for autonomous driving, which present new opportunities to meet all three perception fusion requirements.

Toward this end, we present MMatch, a lightweight perception fusion system that can accurately align the views and localize the vehicles with mmWave radar PCDs. The foundation of MMatch is that fine-grained spatial information, such as the position and angle, provided by the radar can construct unique association with all the vehicles even in the separate views. As a result, if one can capture and understand their associations, including the structural properties between the vehicle and neighbors, its local role played in the structure, and its global position in the view, we can uniquely identify each target and find out the co-visible vehicles. Accordingly, we calculate the transformation and align the views by exploiting the PCD pairs from the co-visible targets.

\begin{figure}[t]
	\centering
	\includegraphics[width=1\linewidth]{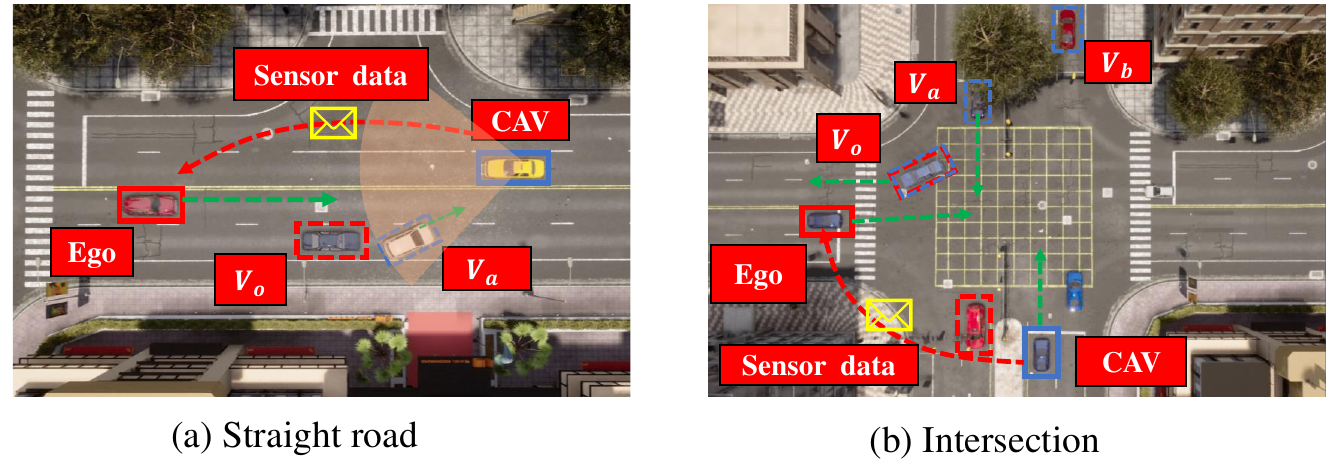} 
	\caption{Advantages of perception fusion in various traffic situations. }
	\label{fig:advantage}\vspace{-0.2cm}
\end{figure}
However, the inherent sparse nature of radar PCDs hinders translating the above idea into reality, where we need to address the following challenges.

\textit{1) How to separate reliable and clear PCDs for alignment from the sparse and noise-polluted PCDs?} Due to specularity and multi-path interference, directly using sparse and polluted PCDs for alignment may bring severe mismatch and high latency, which is inapplicable in the dynamic and complex autonomous driving scenarios. Inspired by the observation that radar PCDs from the stationary and the moving targets present diverse patterns in the radial velocities, we first design a velocity-assistant selection scheme to isolate efficient and clear PCDs of the moving targets. Then, different from recent radar-vision fusion that requires retraining with a large labeled dataset and fine-grained feature extraction on PCDs, we further investigate the propagation patterns between radar and camera, and design a lightweight frustum-based separation scheme by cooperating a widely available monocular camera to accurately filter and separate the radar PCDs of the moving vehicles.

\textit{2) How to construct the association and accurately align the views from sparse PCDs that only cover part areas of the vehicle’s edge.} Different from the high-density LiDAR, sparse radar PCDs are distributed on the vehicle’s edge and lack complete structure, which cannot be exploited to identify co-visible PCD pairs by directly comparing the similarity in complex traffic.  To this end, we first create a graph to construct associations for the targets by using the spatial information of PCDs.  Then, we further design a novel \textit{space-across MPNN} based \textit{RM-net} to pass the message throughout the targets and understand the vehicles’ associations across separate views, which helps to quickly distinguish all the co-visible PCD pairs for alignment. In addition, inspired by the fact that radar PCDs of background provide distinct structural characteristics, we further design a background-constrained alignment algorithm to handle the impact of sparse radar covering on vehicles, which significantly improves alignment accuracy.

\textbf{ Summary of results.} We implement MMatch on both datasets collected from the CARLA and real-world traffic on the campus, and evaluate the performance with over 15,000 radar PCD pairs under various traffic conditions and road types, including straight roads, intersections, and T-junctions. Experimental results show that MMatch can achieve localization errors of 0.7m and 0.9m on the CARLA and real-world datasets, respectively. In addition, MMatch performs the alignment pipeline within 59ms on both two datasets. 

\textbf{Contribution.} First, we present MMatch, the first radar PCD alignment system that realizes accurate and real-time V2V perception fusion with sparse PCDs in the challenging autonomous driving. Second, we design velocity-assistant selection and lightweight frustum-based separation schemes to isolate reliable and clear PCDs for alignment, alleviating the impact of sparsity and interference. In addition, we present a space-across MPNN based \textit{RM-net} to construct and understand the target association, and also a background-constrained alignment approach to handle the impact of sparsity, which significantly reduces time latency and improves alignment accuracy. Finally, we implement MMatch on both CARLA and real-traffic datasets to show the high efficiency of our system.

\begin{figure*}[t]
	\centering
	\includegraphics[width=1\linewidth]{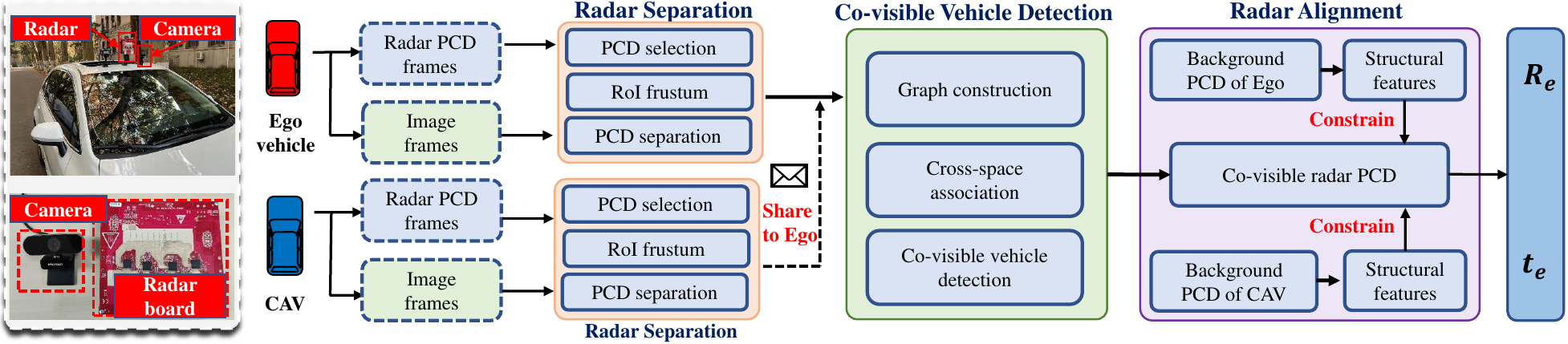} 
	\caption{An overview of our system design. In the first step, the Ego vehicle and CAV perform the operations, separately. After that, CAV shares the image features and radar PCDs to Ego vehicle and finishes the following two steps in the Ego end. }
	\label{fig:overview}\vspace{-0.3cm}
\end{figure*}
\begin{table}[t]
	\centering
	\caption{Limitation of recent alignment approaches.}
	\begin{threeparttable}
		\begin{tabularx}{0.48\textwidth}{p{2.05cm}<{\centering}cccc}
			\toprule
			{\textbf{System }} & \textbf{Accuracy}\tnote{[1]}  & \textbf{Latency}\tnote{[1]} & \textbf{Adaptability}& \textbf{V\&P}  \\ 
			\midrule
			GNSS-based~\cite{shuai2021millieye} & \XSolidBrush  & \Checkmark & \XSolidBrush & {$/$} \\ 
			Feature-level~\cite{he2022automatch}  & \Checkmark  & \Checkmark  & \XSolidBrush & {\XSolidBrush}\\
			Point-level~\cite{he2021vi}  & \Checkmark & \XSolidBrush &  \XSolidBrush & {\XSolidBrush}\\
			
			Object-level~\cite{shi2022vips}  & \Checkmark & \Checkmark  & \XSolidBrush & {\XSolidBrush}\\ 
			\textbf{MMatch } & \Checkmark & \Checkmark  & \Checkmark & {\Checkmark}\\ 
			\bottomrule
		\end{tabularx}
		\begin{tablenotes}    
			\footnotesize              
			\item[{[1]}] \Checkmark in \textbf{Accuracy} indicates that the system achieves decimeter-level accuracy, and \Checkmark in \textbf{Latency} indicates time consumption is within 100ms.                   
		\end{tablenotes}           
		\label{table:limitation}
	\end{threeparttable} 
\end{table}
\section{Motivation} \label{sec:motivation} 
To show benefit of perception fusion, we present typical driving examples in CARLA. Specifically, we deploy two cooperative vehicles (e.g., Ego and CAV) in two typical traffic scenarios, including the straight road and the crossroad. As shown in Fig.~\ref{fig:advantage}(a), due to the occlusion by a large vehicle $V_o$, Ego vehicle fails to detect the vehicle $V_a$ that is performing lane change, which may lead to traffic collision.  However, the Ego vehicle can avoid this issue with shared results from the CAV. Similarly, as presented in Fig.~\ref{fig:advantage}(b), Ego vehicle can see through the barriers and capture more targets in a crowded crossroad by aligning the detection results from CAV, which is helpful to generate safe planning with the perception fusion.

Recent advances have designed a bunch of approaches to align the views, including the GNSS-based scheme, the image feature-level approach, and the raw data point-level and object-level LiDAR systems. As presented in Table~\ref{table:limitation},  we first investigate GNSS-based scheme~\cite{arnold2020cooperative}, where it achieves an average localization error about 5.6m in 55ms by the impact of measurement errors.  In the image feature-level system, we extract landmark keypoints of the ground signs~\cite{he2022automatch}, and have decimeter-level (0.3m) localization errors and latency within 100ms (90ms) in open spaces. While in crowded traffic, it fails to detect the ground sign and align the views. Then, we investigate the performance in LiDAR system. In point-level experiments~\cite{he2021vi}, we also achieve decimeter-level (0.39m) localization error while bringing the latency higher than 300ms. In the object-level system, we share detected targets from the LiDAR to fuse the views~\cite{shi2022vips} which achieves decimeter-level (0.6m) accuracy within 100ms (63ms) in light traffic. However, if the targets stay close and have similar geometries in a crowded street, this approach is hard to distinguish the co-visible agents. Despite high accuracy, all this density-LiDAR based view alignment approaches fail to applied to radar system due to the PCD sparsity. In addtion, among all these sensors, only radar PCDs can provide both velocity (\textbf{V}) and position (\textbf{P}) information in a single frame, which are crucial in autonomous driving. Therefore, perception fusion with radar is important while challenging in autonomous driving.

\section{System Design} \label{sec:design} 
\subsection{System Overview}\label{sec: ove} 
As illustrated in Fig.~\ref{fig:overview}, the pipeline for MMatch consists of  the following three key steps:
\begin{itemize}
	\item \textbf{Radar separation.} Due to the radar PCD sparsity and interference, our first step requires to separate reliable and efficient PCDs for alignment. To this end, we design a lightweight PCD selection and separation scheme by cooperating the radar and camera. 
	
	\item \textbf{Co-visible vehicle detection.} With the image-radar frames shared from the CAV, we design a space-across MPNN based \textit{RM-net} to construct the association among the vehicles and understand its unique position for the targets, from which MMatch can accurately lock the co-visible vehicles of two separate views.
	
	\item \textbf{Radar alignment.} To avoid the radar PCD mismatch caused by the large different view angles as well as radar sparsity, we further design a background-constrained alignment approach, which reduces the co-visible vehicle PCD alignment errors by fully using the structural features of the stationary background.
\end{itemize}

\subsection{Radar Separation}\label{sec: pre} 
Due to the specular reflection and interference, simply leveraging all the PCDs for view alignment will introduce severe mismatch and complex computation. To this end, we first design a velocity-assistant selection scheme to acquire efficient and clear PCDs of the moving objects. Then, different from recent radar-vision fusion approaches that requires retrain with large labeled dataset and fine-grained feature extraction, we cooperate a monocular camera and design a lightweight frustum-based separation scheme to accurately filter and separate the radar PCDs of the moving vehicles. 

\begin{figure}[t]
	\centering
	\includegraphics[width=1\linewidth]{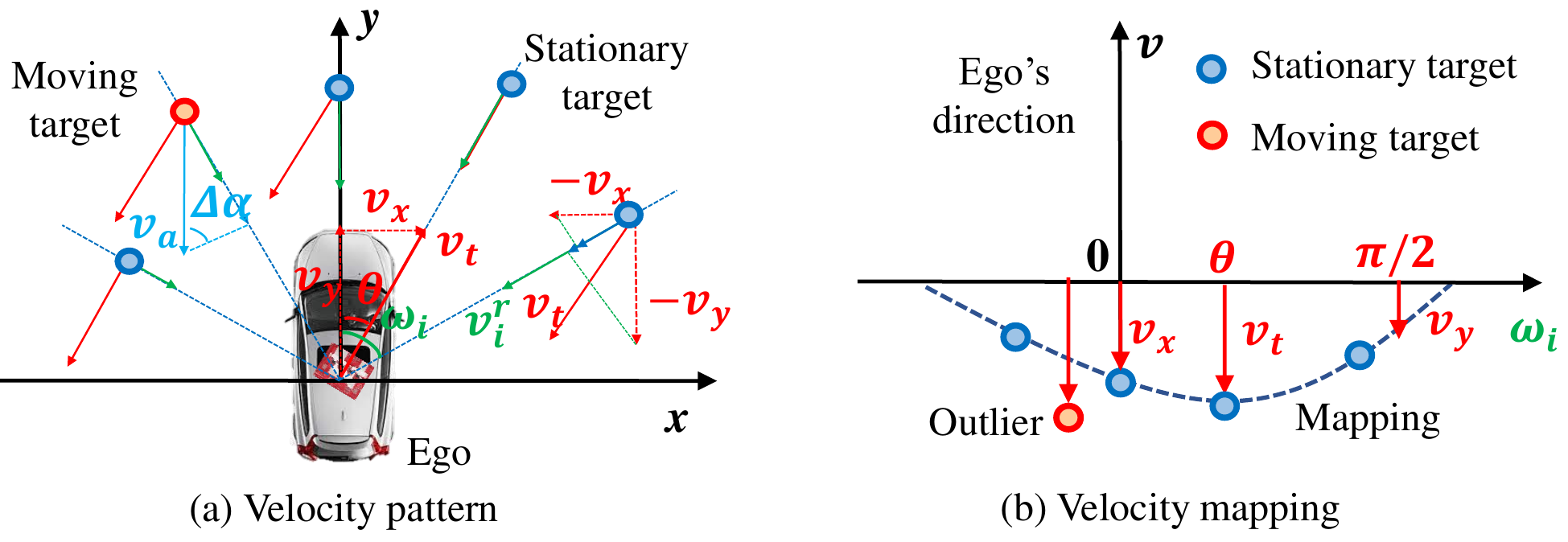} 
	\caption{PCD selection. (a) Moving and stationary targets show different speed patterns. (b) PCDs from stationary points are mapped to a sinusoidal curve, while the points from moving targets appear as outliers.
	}
	\label{fig:radar_selection}\vspace{-0.3cm}
\end{figure}

\begin{figure}[t]
	\subfigure[Radar point mapping.]
	{\includegraphics[width=1.68in]{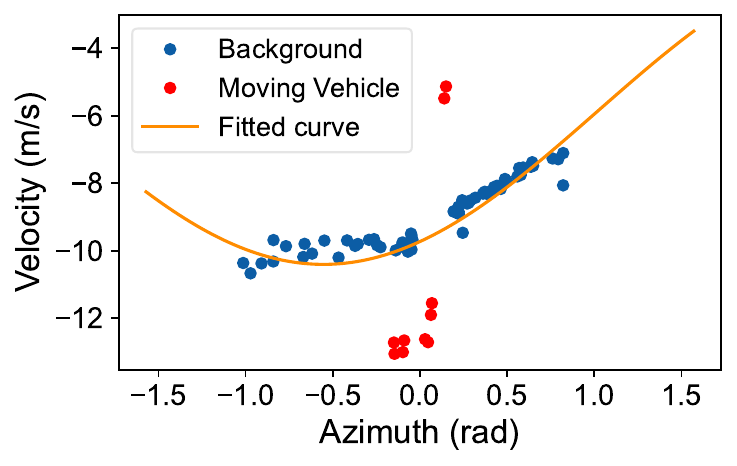}}
	\subfigure[Radar PCD separation.]
	{\includegraphics[width=1.8in]{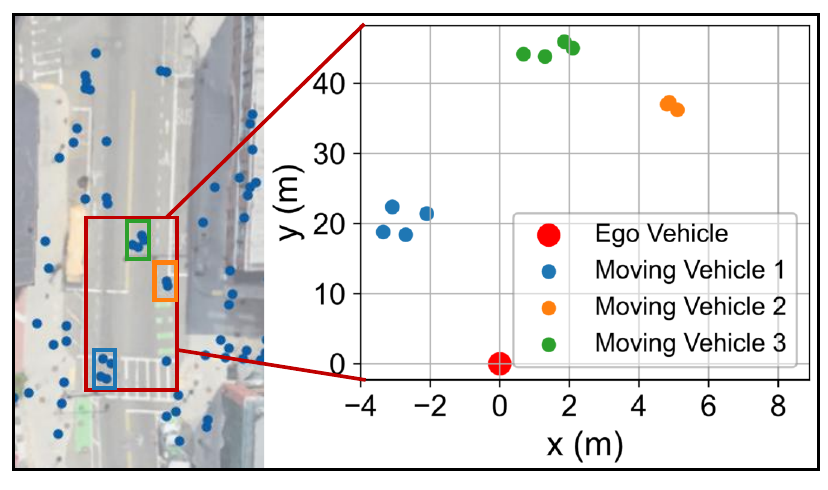}}
	\caption{We isolate the radar PCDs of the moving targets from the background.}
	\label{fig:radar_separation_example}\vspace{-0.3cm}
\end{figure}

\textbf{PCD selection.} We first employ a lightweight velocity-assistant selection scheme to isolate radar PCDs of the moving targets. Our insight is that radar PCDs of the stationary background and moving objects present significant different patterns in radial velocities~\cite{kellner2013instantaneous,gao2022dc}, from which we can identify the PCDs of the moving objects. Specifically, as shown in Fig.~\ref{fig:radar_selection}(a), Ego vehicle is moving from the Y-axis direction, where a radar is attached with a deviation angle of $\theta$ from the vehicle's heading direction. Since Doppler radar captures only the radial velocities, all the stationary objects have the same velocity $v_t$  but an opposite direction to the radar module. Accordingly, we can project the stationary targets' velocity $v_t$ to the X-axis and Y-axis directions, and achieve the velocity components as $v_x=-v_t\sin(\theta)$ and $v_y=-v_t\cos(\theta)$, respectively. At the same time, any of the point $i$ with an azimuth position angle $\omega_i$, we acquire its radial velocity as
\begin{equation}\label{eq:pro}
	\begin{aligned}
		v^r_i = -v_t\sin(\theta)\sin(\omega_i)-v_t\cos(\theta)\cos(\omega_i).
	\end{aligned}
\end{equation}
According to Eq.~(\ref{eq:pro}) and Fig.~\ref{fig:radar_selection}(b), all the stationary radar points fall on a sinusoidal curve. In contrast,  for a moving vehicle with a velocity of $\Delta \alpha$, we have its radial velocity as
\begin{equation}\label{eq:prop}
	v^r_i =v_a\sin(\Delta \alpha)- v_t\sin(\theta)\sin(\omega_i)-v_t\cos(\theta)\cos(\omega_i).
\end{equation}
Compared with Eq.~(\ref{eq:pro}) and Eq.~(\ref{eq:prop}), radar points coming from a moving vehicle will be out of the sinusoidal curve as outliers due to an additional term of its movement. As an example of the traffic in Fig.~\ref{fig:radar_separation_example}(a), we successfully identify the radar PCDs of the moving targets.

\begin{figure}[t]
	\centering
	\includegraphics[width=1\linewidth]{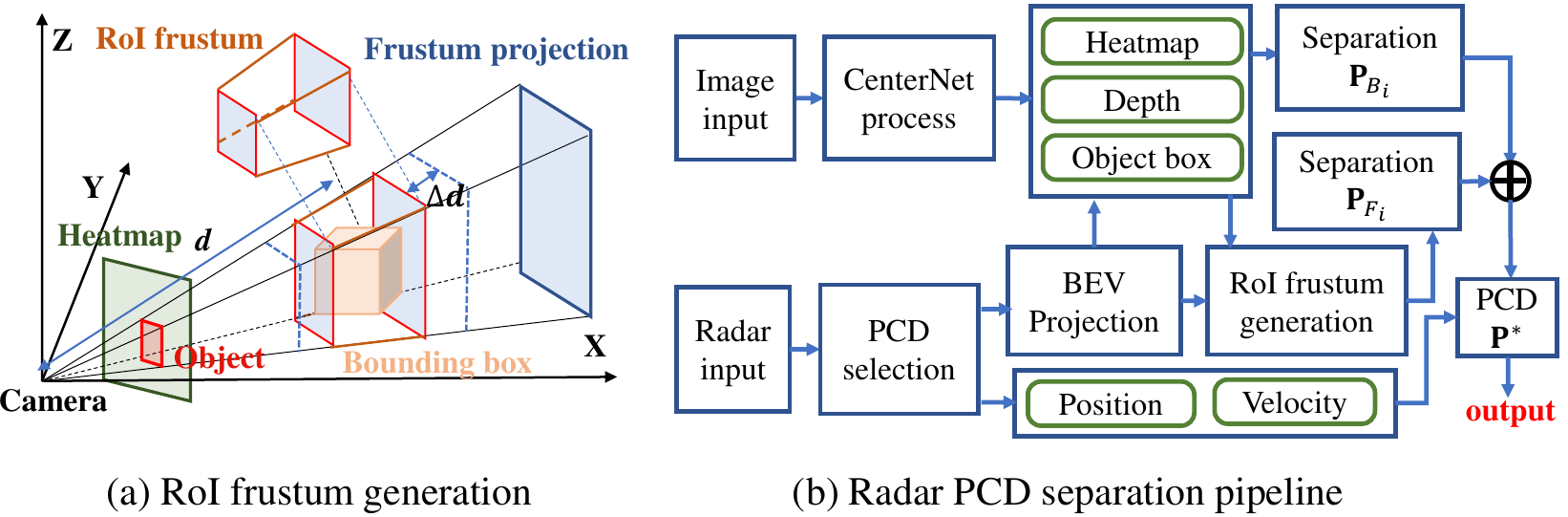} 
	\caption{PCD separation.}
	\label{fig:radar_separation}\vspace{-0.3cm}
\end{figure}
\textbf{Frustum-based separation.} With the PCDs of the moving target, we cooperate a monocular camera to accurate separate and map them to the vehicles. To achieve 3D detection from a 2D image, we first leverage the well-trained CenterNet~\cite{zhou2019objects}  to detect the vehicles directly, where it takes images as input and output a series of 3D bounding boxes, a 2D heatmap, and the depths to indicate the vehicles. In each box $B_i$, one may project the radar PCDs to the box for data separation as $\textbf{P}_{B_i}$. However, CenterNet brings bounding box estimation errors due to the lack of real depth information, resulting in false projection and separation of the PCDs. Inspired by~\cite{nabati2021centerfusion}, we further employ a frustum-based separation scheme to isolate the PCDs. The basic idea is that both the radar and camera present a frustum propagation pattern, which can be cooperated to determine the PCD distributions. Specifically, as shown in Fig.~\ref{fig:radar_separation}, we create a frustum propagation with a 2D heatmap $\textbf{I}(u,v)=\textbf{K}[X, Y, Z]^T\in\mathbb{R}^{W\times H}$, where $W$ and $H$ represent the width and height,  $\textbf{K}$ is the camera intrinsic matrix, and $\textbf{L}=[X, Y, Z]$ denote the position in the frustum. Then, combined with the object box and its estimated depth $d=\|\textbf{L}\|$, we can determine a RoI frustum to indicate the possible distribution space of the PCDs. To alleviate the impact of depth estimation inaccuracy, we enlarge the RoI frustum with a depth of $\Delta d$ to including more adjacent PCDs. In addiction, measurement errors in height dimension may result in false separation of the PCDs. To this end, we project the PCDs into the BEV and determine the projected points inside the RoI frustum $F_i$ corresponding to this object, which can be represented as $\textbf{P}_{F_i}$. Finally, we can collect a series of PCDs $\textbf{P}^*=\textbf{P}_{B_i}\cup \textbf{P}_{F_i}$ corresponding to this moving vehicle $i$.

Instead of labeling a large dataset to train and correct the PCD separation as~\cite{nabati2021centerfusion}, we design a lightweight spatial-Doppler filtering approach to filter out the noise. Our insight is that the radar points reflected by the same target would get closer and present in the form of clusters, while the points polluted by the noise may scatter with a random distribution. Moreover, the radar points from the same vehicle will show similar velocities and positions, making the points present a clustering pattern in both the velocity and position dimensions. Therefore, we take both the positions and velocities of the PCDs as inputs for constrain and use a DBSCAN-based filter $\mathcal{F}(\textbf{P}^*)$ to sanitize the PCDs. As shown in Fig.~\ref{fig:radar_separation_example}(b), we separate vehicle radar points from the stationary background and remove the unreliable points polluted by the noise.

\textbf{Data transmission.} After separation, a CAV  transmits the image-radar frames to Ego vehicle for perception fusion. Thanks to sparse nature of PCDs, CAV transmits both PCDs of the moving vehicles and also the stationary background to improve accuracy. As for image frames, since the raw data will occupy a large bandwidth, we compress the detection vehicles for the heatmap with recent \textit{Mobilenet}v2~\cite{sandler2018mobilenetv2} and share only feature maps of the detected vehicles by CenterNet.

\subsection{Co-visible Vehicle Detection}\label{sec: cov} 
After receiving the radar-image frames, the Ego vehicle tends to distinguish the co-visible vehicles for view alignment. Recent LiDAR-based solution~\cite{shi2022vips} compares the geometry similarity among the detected objects to identify the co-visible targets, which however poses several challenges for the sparse radar. First, it relies on accurate velocity and relative complete object structures to define their relationships, while sparse radar PCDs are not enough to accurately estimate the heading speed and complete structures. Second, the geometry constructed by similar vehicles from some local areas would also present similar structures, resulting in false co-visible target recognition in a crowded and complex traffic. 

To overcome this predicament, we design a space-across MPNN based \textit{RM-net} to learn and understand the vehicles’ association, which then help to uniquely identify the co-visible vehicles across the views. Specifically, we first construct a graph where the nodes correspond to the semantic features from images, and the edges represent connections using spatial information from PCDs. Then, to bridge the gap of separate views, we design a space-across MPNN to update the nodes and edges, and understand the vehicles’ association. Finally, we define a multi-label classifier to weigh the association and identify the co-visible vehicles.

\begin{figure}[t]
	\centering
	\includegraphics[width=1\linewidth]{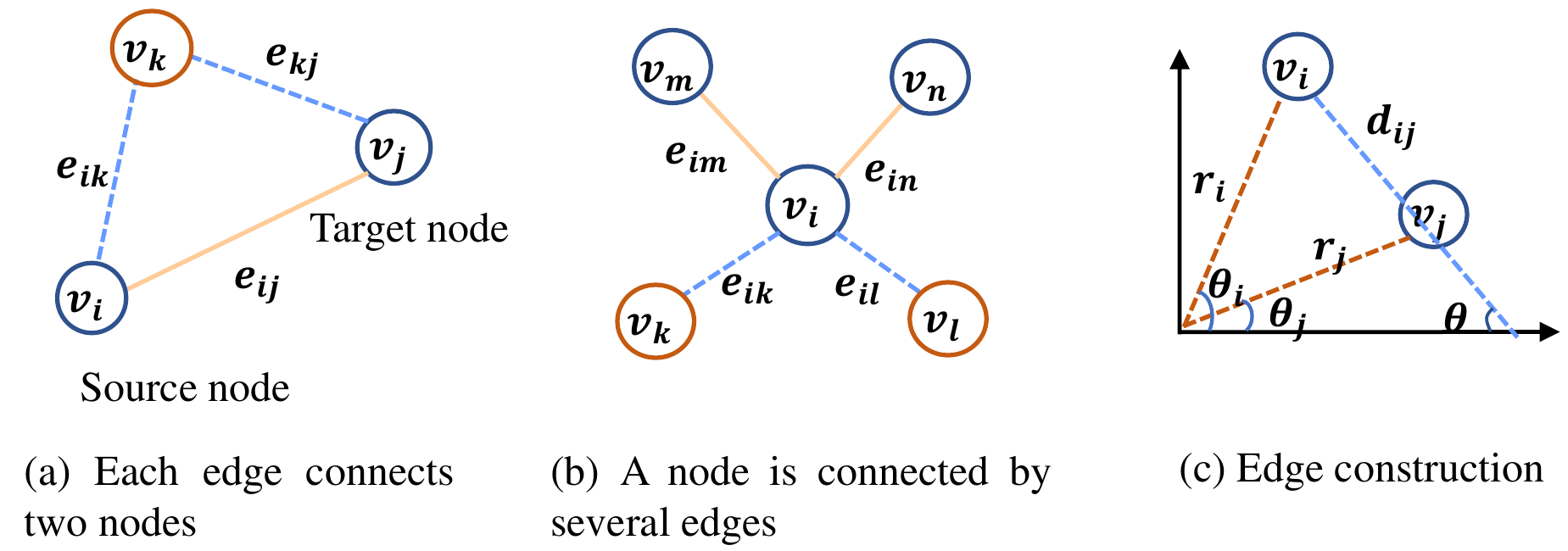} 
	\caption{Graph construction. }
	\label{fig:graph_construction}\vspace{-0.2cm}
\end{figure}

\textbf{Graph construction.} As illustrated in Fig.~\ref{fig:graph_construction}, our first step is to construct an undirected graph $\mathbf{G}=(\mathbf{V}, \mathbf{E})$ with image-radar frames. In our graph, the nodes ${ v_i}\in\mathbf{V}$ correspond to the detected vehicles from the images, while the edge $e_{ij}\in\mathbf{E}$ represents the possible association by the radar PCDs. Recent study has proved that low-dimensional vector embeddings are high-efficient in a large graph~\cite{cao2015grarep}. Therefore, we exploit a encoder-based dimension reducer to standardize the high-dimensional image features and irregular radar PCDs to unified vector embeddings. Specifically, in the node end, we directly extract a feature embedding using a MLP-based encoder $\mathcal{E}_v$, where the corresponding initial node embedding for node $i$ is ${m}^{(0)}_{v_i}=\mathcal{E}_v(v_i)$. As for the edge end, due to different reference systems, we separate the edge determination into \textit{local-vehicle construction} and \textit{across-vehicle construction} stages. As shown in Fig.~\ref{fig:graph_construction}(c), in the \textit{local-vehicle construction} stage, we seek to connect the nodes (the solid lines in Fig.~\ref{fig:graph_construction}(a)) by relying on the related distance and angle as 
\begin{equation}
	e_{ij}=\left( {d_{ij}\over d_{max}},{\theta\over{\pi}},{\log{r_i\over{r_j}}},{\log{\theta_i\over{\theta_j}}}\right)
\end{equation}
where $d_{ij}$ denotes the distance between a pair of nodes $i$ and$j$, $d_{max}$ represents the maximum of all the edges, $\theta_i$ and $\theta_j$ are the angles from the sensor's view, and $\theta$ is the angle of the edge $e_{ij}$. Then, in the \textit{across-vehicle construction} stage, we assign an initial value of 0 for the nodes (the dot lines Fig.~\ref{fig:graph_construction}(a)). Finally, we feed the edge into an MLP-based encoder $\mathcal{E}_e$ to obtain the initial edge embedding as ${m}^{(0)}_{e_{ij}}=\mathcal{E}_e(e_{ij})$.

In embedding processing, image feature for each vehicle is distilled into a 1$\times$32 node embedding while the corresponding radar PCD is encoded to a 1$\times$16 edge embedding. After that, we construct the connections for all the targets and learn the association of all the targets  to determine the co-visible vehicles by weighing the dot lines across the views.

\begin{figure*}[t]
	\centering
	\includegraphics[width=1\linewidth]{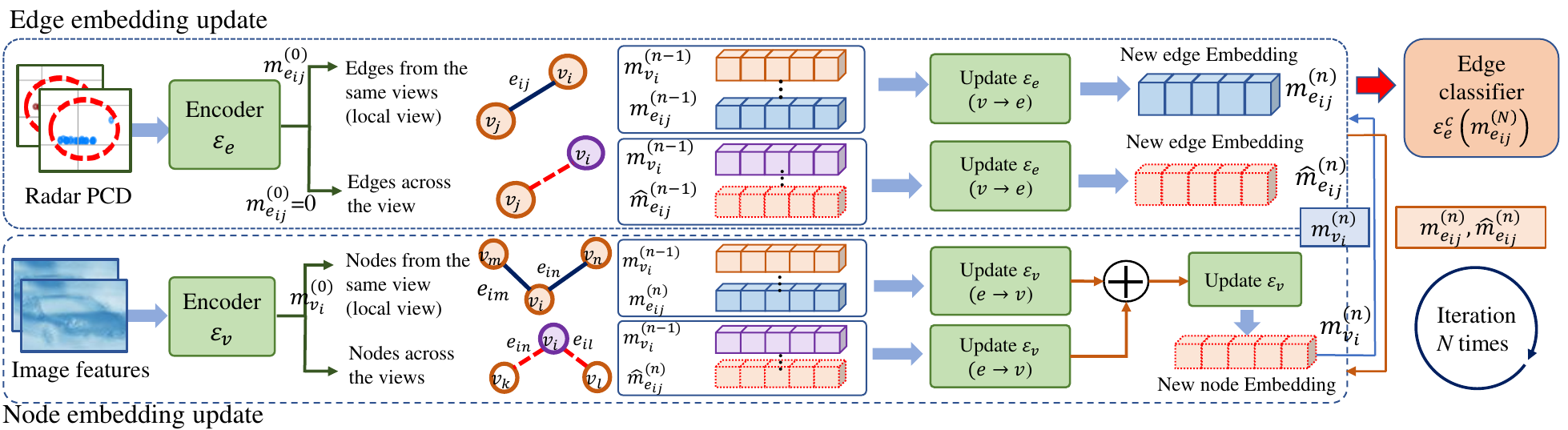} 
	\caption{RM-net uses a space-across MPNN to learn and update the association of the all targets in the graph. }
	\label{fig:space_MPNN}
\end{figure*}
\textbf{Space-across MPNN.} 
Given a constructed graph, an efficient solution to link the association can employ MPNN to propagate the information across the node and edge~\cite{gilmer2017neural,tedeschini2022addressing}. However, unlike the consecutive-frame system in a single agent~\cite{braso2020learning}, the nodes and edges in cooperative vehicle system cannot directly update the embeddings across two separate spaces. Towards this end, as shown in Fig.~\ref{fig:space_MPNN}, we design a space-across MPNN for embedding update, where we perform \textit{edge embedding update} and \textit{node embedding update} in both the local view and across views.

\textit{Edge embedding update.} As shown in Fig.~\ref{fig:graph_construction}(a), each edge $e_{ij}\in \mathbf{E}$ connects two node embeddings, where we denote them as source and target node embeddings, respectively. Therefore, the edge embedding update requires taking into account both the source, target and edge embeddings. Since the nodes from the Ego vehicle and cooperative CAV share separate space, we divide the edges into two classes, one connecting the nodes in the local view and the other connecting vehicles across spaces, and then we perform edge embedding update for them. In particular, we assume that \textit{RM-net} performs $N$ maximum message passing steps in the graph. Accordingly, in the local view, the edge embedding update ${m}^{(n)}_{e_{ij}}$ for a pair of nodes at iteration $n$ can be denoted as
\begin{equation}
	(v\rightarrow e): {m}^{(n)}_{e_{ij}}={\mathcal{E}_e} \left( [{{m}^{{(n-1)}}_{v_i}}, {{m}^{(n-1)}_{v_j}}, {{m}^{(n-1)}_{e_{ij}}} ]  \right).
\end{equation}
As for the edges across views, we can update the edge embeddings by performing similar operations, where we denote the updated edge embeddings at the iteration $n$ as $\hat{m}^{(n)}_{e_{ij}}$. As a result, we can traverse all the edges in the graph and achieve the edge embeddings as ${m}^{(n)}_{E}=\{ {m}^{(n)}_{e_{1}},...,{m}^{(n)}_{e_{p}},..., \hat{m}^{(n)}_{e_{1}},...\hat{m}^{(n)}_{e_{q}}\}$.

\textit{Node embedding update.} As shown in Fig.~\ref{fig:graph_construction}(b), each node $v_i\in\mathbf{V}$ is connected by a set of edges, and thus the node embedding update needs to iterate both the connecting edges and node embedding itself. Similar to the edge embedding update, we also perform the update operations in both the local view and across views. In particular, we first update the edge embeddings ${d}^{(n)}_{e_{ij}}$ in the local view as 
\begin{equation}
	(e\rightarrow v): {d}^{(n)}_{e_{ij}}={\mathcal{E}_v} \left( [{{m}^{(n-1)}_{v_i}},  {{h}^{(n)}_{e_{ij}}} ]  \right)
\end{equation}
where ${{m}^{(n-1)}_{v_i}}$ denotes the last node and ${{h}^{(n)}_{e_{ij}}}\in{{m}^{(n)}_{E}} $ is the current edge embedding. Following this update rule, we achieve the edge embeddings across the view as $\hat{d}^{(n)}_{e_{ij}}$. Finally, we can update the node embedding ${m}^{(n)}_{v_i}$ and connect two spaces by combining these two edge embedding classes as
\begin{equation}
	{m}^{(n)}_{v_i}={\mathcal{E}_v} \left( \sum_{e_{ij}\in{{d}^{(n)}_{e_{ij}}} }  {d}^{(n)}_{e_{ij}} \oplus \sum_{e_{ij}\in{\hat{d}^{(n)}_{e_{ij}}} } \hat{d}^{(n)}_{e_{ij}}  \right)
\end{equation}

\textbf{Training and loss function}. After $N$ message passing iterations, all the nodes have shared information for each other in the entire graph. Our goal is to find out the co-visible targets from a series of edges $\hat{e}_{ij}$ that connect the nodes from two separate views. To this end, we feed all the output edge embeddings ${m}^{(N)}_{E}$  into an MLP-based classifier $\mathcal{E}^{class}_{e_{ij}}$, which predicts the association probabilities $\hat{y}^{(N)}_{e_{ij}}$ for every edge $e_{ij}\in \mathbf{E}$. In this classifier, it determines whether an edge associates the same target from two different views by performing a comparison  operation with a threshold $\eta$ as
\begin{equation}
	\hat{y}^{(N)}_{e_{ij}}= {\mathcal{E}^{class}_{e_{ij}}} {\left( {{m}^{(N)}_{E}} \right)} \ge \eta           
\end{equation}
In order to train the model, our loss function is defined with a weighted binary cross-entropy for the co-visible vehicle prediction in the last message passing iteration $n$ as
\begin{equation}
	{\mathcal{L}}= { {-1\over{\left|{\mathbf{E}}\right|} } \sum_{n=1} ^N \sum_{{e_{ij}}\in{\mathbf{E}} }  (1-{y_{e_{ij}}})\log(1-{\hat{y}^{(n)}_{e_{ij}}}) +  \omega\cdot{y_{e_{ij}}}\log {y^{(n)}_{e_{ij}}}                   }
\end{equation}
where $\omega $ is a weight to compensate for the high imbalance between the edges of co-visible vehicles and the others, and it is defined as
\begin{equation}
	\omega={   {\sum_{e_{ij}\in{\mathbf{E}}}\mathbf{1}(y_{e_{ij}}=0)} \over{\sum_{e_{ij}\in{\mathbf{E}}}\mathbf{1}(y_{e_{ij}}=1)}                }                
\end{equation}
where $\mathbf{1}(\mathbf{x})$ represents an indicator function that returns the value 1 when the condition $\mathbf{x}$ is satisfied and 0 otherwise. As a result, the system prefers to assign a larger weight to the edge classes that connect the same co-visible vehicles.

\subsection{Radar Alignment }\label{sec: bac}  
After the processing pipeline, we have a series of PCD pairs $\textbf{P}^E_n$ and $\textbf{P}^C_n$corresponding to the co-visible vehicles, where $\textbf{P}^E_n$ and $\textbf{P}^C_n$ indicate the $n_{th}$ PCD pairs from Ego vehicle and the CAV. An intuitive way of estimating the transformation $\textbf{T}=(\textbf{R}_e, \textbf{t}_e)$ is to perform Iterative Closest Point (ICP), where $\textbf{R}_e$ and $\textbf{t}_e$ denote the rotation and translation matrix, respectively. However, sparse PCDs and large different views pose two challenges in view alignment. First, we may achieve only limited PCD pairs if only several co-visible vehicles are in the view. Second, large differences in viewing angles between the Ego vehicle and CAV result in significantly different coverage areas even for the same target. For example, PCDs from Ego vehicle and CAV cover the front and the rear of a vehicle, respectively. In this case, simply finding the closest points by ICP may bring severe alignment errors.

To this end, we further design a background-constrained alignment algorithm to improve accuracy. Our key insight is that the PCDs collected from roadsides and other stationary object present distinct structural features, which can be harnessed to constrain the alignment. However, typical road types, such as straight roads and intersections, present perfect axisymmetric and rotational symmetry, which would mislead the alignment to an opposite or rotational symmetry direction. Therefore, directly combining radar point pairs of the co-visible vehicle and the background may bring a negative impact on the alignment. To this end, we first employ DBSCAN to filter the outlier caused by the noise and multi-path reflection, preserving the point clusters of the roadsides and buildings. Then, instead of performing ICP by directly combining the PCDs, we first conduct iterations only on the PCD pairs of the co-visible vehicles to obtain an initial transformation $\textbf{T}_0$. The iteration period is defined as $\min\{L, N(D_{max})\}$, where $L$ represents half of the total number of iterations of the co-visible vehicle radar point pairs, and $N(D_{max})$ is the number of iterations that achieves the convergence distance threshold $D_{max}$. Since the spatial distribution of the co-visible vehicles poses well geometry, we can achieve an approximate rotation $\textbf{R}_0$ to avoid the alignment error caused by roadside symmetry. Finally, we continue the iterations by combining both the co-visible PCD pairs of $\textbf{P}^E_n$ and $\textbf{P}^C_n$, and the background PCDs of $\textbf{B}^E$ and $\textbf{B}^C$, where we design optimization rule as
\begin{equation}
	\textbf{T}={ \mathop{\arg\min}\limits_{\textbf{T} } \{\sum_{n}^N\omega_n\mathbf{E}_i(\textbf{P}^E_n,\textbf{P}^C_n)+\omega_g\mathbf{E}_g(\textbf{B}^E,\textbf{B}^C)\}}                       
\end{equation}
where $\mathbf{E}_n$ and $\mathbf{E}_g$ denote the estimation errors caused by co-visible PCD pairs and background PCDs. $\omega_n=e^{score_n}$ and $\omega_g=e^{1\over{N_v}}$ are the weights assigned to them. The insight is that, if vehicle detection achieves a high score by CenterNet, it assigns a higher weight. While if we detect only a small number of the co-visible vehicles $N_v$, we will rely more on the background radar and assign a higher weight to it.
\begin{figure}[t]
	\centering
	\includegraphics[width=0.9\linewidth]{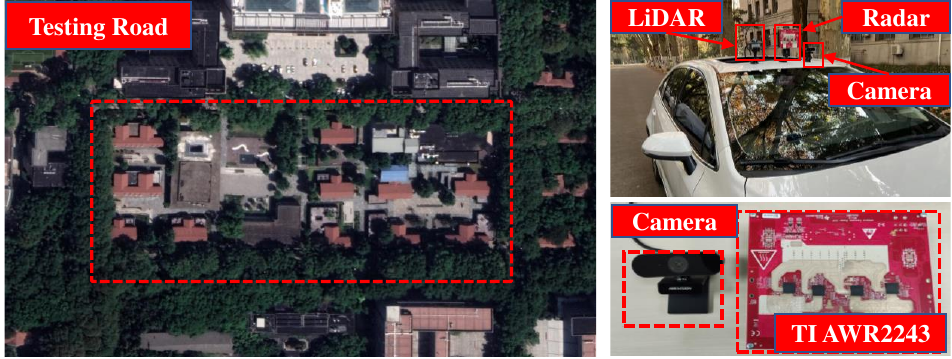} 
	\caption{Data collection in the real-world traffic on the campus. }
	\label{fig:imple}\vspace{-0.2cm}
\end{figure}

\section{Implementation}\label{sec:implementation} 
Since there are no publicly available radar PCD dataset for cooperative vehicle perception fusion, we collect datasets on both the CARLA platform and real-world deployments. 

\textbf{CARLA dataset.} We first create dataset using CARLA simulator for performance comparison without the impact of noise.  In our experiments, we drive connected vehicles through city and town traffic, navigating a variety of typical road types, like straight roads, intersections, and T-junctions. Similar to~\cite{shi2022vips}, we enable the vehicles to collect data at diverse traffic, including the light traffic (less than 5 vehicles) and heavy traffic (more than 5 vehicles). In order to evaluate the system robustness, we also vary driving modes of the cooperative vehicles, including driving from the same, the opposite, and vertical directions, where the traffic conditions experience different illuminations. In our system, vehicles are set to a speed of 8-20m/s. In this dataset, we collected over 2,000 image-radar frames containing over 10,000 target pairs, where 1,000 frames are used to train the model and the rest are for performance testing.

\textbf{Real-world dataset.} We also implement our system in real-world traffic scenarios. As shown in Fig.~\ref{fig:imple}, we first install a pair of monocular cameras and TI AWR2243 radar boards on both the Ego vehicle and CAV. Then, we drive the vehicles with a velocity of 5.5-8.3m/s on the campus to collect real-world data in both the straight roads, T-junctions, and intersections. In order to obtain the ground-truth transformation, we construct HD maps for all the experimental roads. Specifically, we first deploy a LIVOX LiDAR board to repeatedly scan the streets and construct HD maps by~\cite{9372856}. Then, by aligning vehicles' LiDAR with the HD map, we achieve the ground-truth transformation between the Ego vehicle and CAV. In our experiments, we have collected over 1000 image-radar frames (containing over 3,000 vehicle PCD pairs) that also experience diverse illumination as well as dust and smog conditions.

\begin{figure}[t]
	\subfigure[RRE estimation.]
	{\includegraphics[width=1.67in]{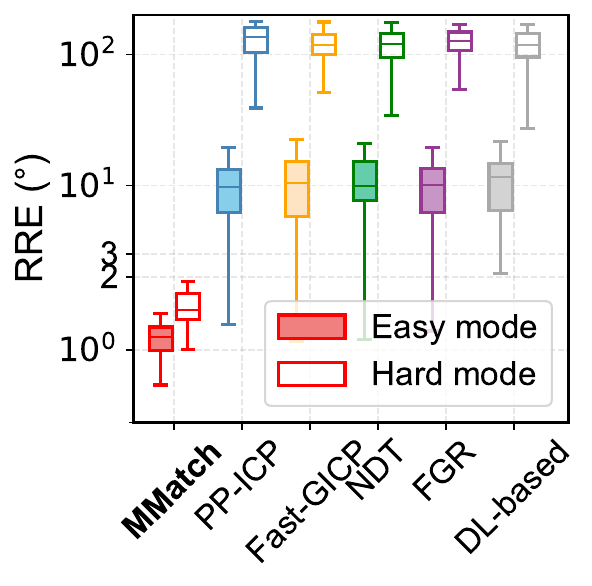}}
	\subfigure[RTE estimation.]
	{\includegraphics[width=1.67in]{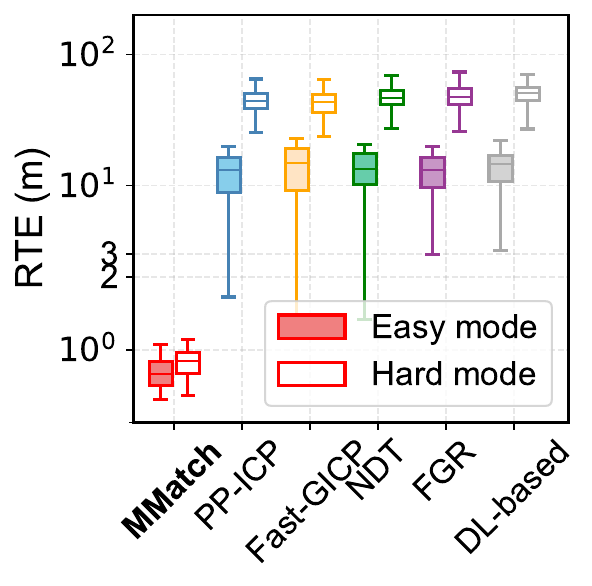}}
	\caption{Performance of alignment accuracy on CARLA dataset.}
	\label{fig:aligment_acc}\vspace{-0.3cm}
\end{figure}
\textbf{Experiment setup.} We implement MMatch on the PCs, each equipped with a Intel(R) Xeon(R) Gold 6226R 2.90GHz CPU and NVIDIA GeForce RTX 3060ti GPU. To compress the image, we use pre-trained \textit{MobileNet}v2 to extract the features for all the image frame inputs~\cite{sandler2018mobilenetv2}. To evaluate the effectiveness of the dynamic channel, cooperative vehicles carries a pair of 802.11ac-based WiFi routers that share image-radar frames in a variety of traffic conditions, including open areas and roads that are densely obscured by trees. In the model training stage, we train our \textit{RM-net} model on the GPU, while the performance testing is working only with the CPU.

\section{Evaluation}\label{sec:evaluation} 
\subsection{Evaluation Metrics} 
In our experiment, we define the following metrics to evaluate the performance of MMatch, including,
\begin{itemize}
	\item \textbf{RRE.} Relative rotation error (RRE) is to measure the bias between the real rotation $R_t$ and estimated value $R_e$, and it is defined as $RRE=\left\|{f}(R^{-1}_tR_e)\right\|_1$, where $f$ is the function to calculate the three Euler angles for the rotation matrix.
	\item \textbf{RTE.} Relative translation error (RRE) is to measure the bias between the real translation $t_t$ and estimated value $t_e$,  which is defined as $RRE=\left\|t_t-t_e\right\|_2$.
\end{itemize}
In the following sections, we evaluate the performance on both the CARLA and the real-world datasets.

\begin{figure}[t]
	\subfigure[RRE estimation.]
	{\includegraphics[width=1.67in]{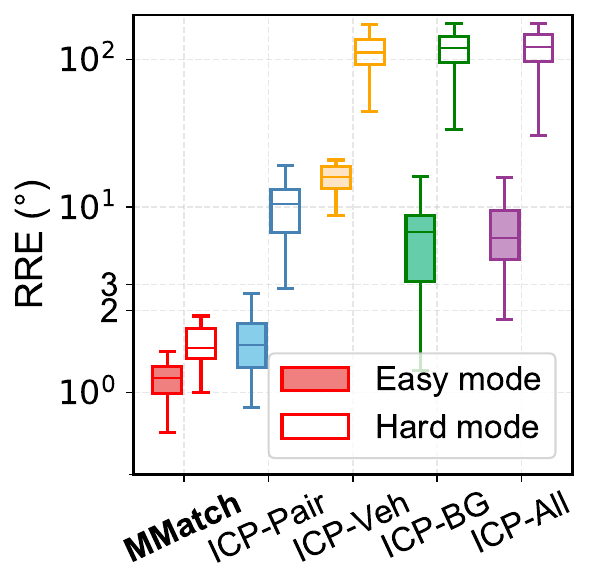}}
	\subfigure[RTE estimation.]
	{\includegraphics[width=1.67in]{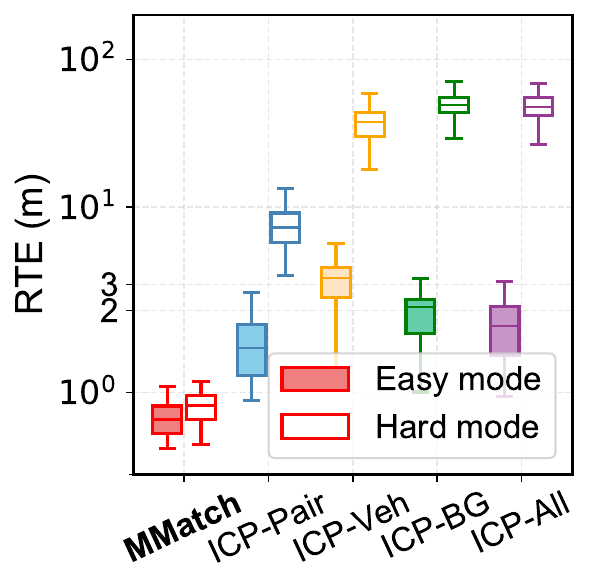}}
	\caption{Performance of ablation experiments on CARLA dataset.}
	\label{fig:ablation_acc}\vspace{-0.3cm}
\end{figure}
\subsection{Alignment Accuracy on CARLA Dataset}\label{sec:ca_accuracy} 
We first evaluate the alignment accuracy on the CARLA dataset, where we divide our experiments into easy and hard modes. Specifically, in the easy mode, we enable the vehicles to drive from the same direction on straight roads and collect data in light traffic. While in the hard mode, we deploy vehicles on diverse road types. Then, we drive the Ego vehicle and CAV from different directions and speeds to collect data in heavy traffic. In the experiments, all vehicles experience narrow lanes (2-4 lanes) and wide lanes (6-10 lanes) and the speed is set as 8-20m/s.

\textbf{Performance comparison.} We compare performance with the representative alignment approaches, including classical PP-ICP~\cite{low2004linear} and NDT~\cite{biber2003normal}, recent  Fast-GICP~\cite{koide2021voxelized} and FGR~\cite{zhou2016fast}, and deep learning based approach DL-based~\cite{ao2023buffer}. As shown in Fig.~\ref{fig:aligment_acc}, all these approaches achieve accuracy in RRE higher than $9^\circ$ and RTE higher than 15m even in the easy mode. The reason is that all these approaches are simply to iteratively find the closest points from the PCDs. However, sparse radar PCDs lack complete structural information and cover only part of the area of a vehicle. As a result, simply iterating to match the points with the shortest distance brings much error to the results. In contrast, MMatch achieves alignment accuracy in RRE about $1.2^\circ$ and RTE about 0.6m in the easy mode, and achieves RRE about $1.6^\circ$ and RTE about 0.8m in the hard mode, which significantly enable many driving tasks, such as overtaking and safe planing.

\begin{figure*}[t]
	\hfill
	\begin{minipage}[t]{0.25\linewidth}
		\centering
		\includegraphics[width=1\linewidth]{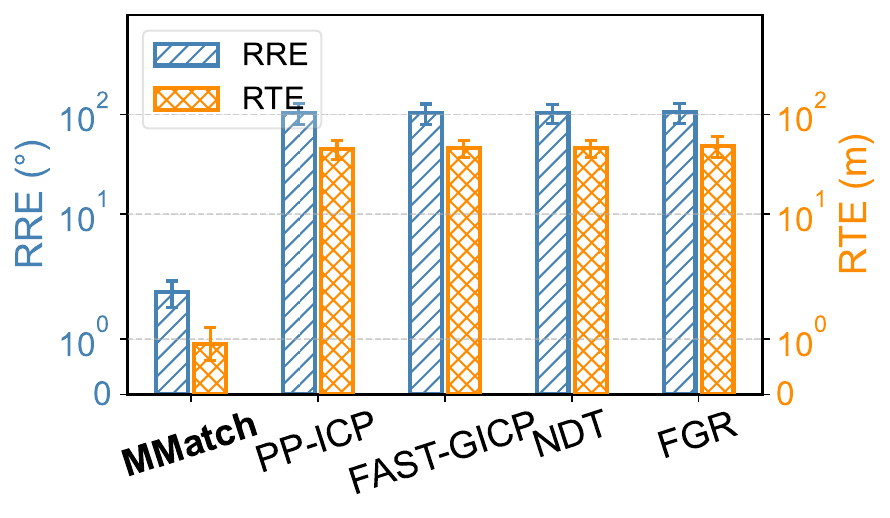}
		\caption{Performance of alignment accuracy on real-world dataset.}
		\label{fig:real_bas}
	\end{minipage}
	\hfill
	\begin{minipage}[t]{0.25\linewidth}
		\centering
		\includegraphics[width=1\linewidth]{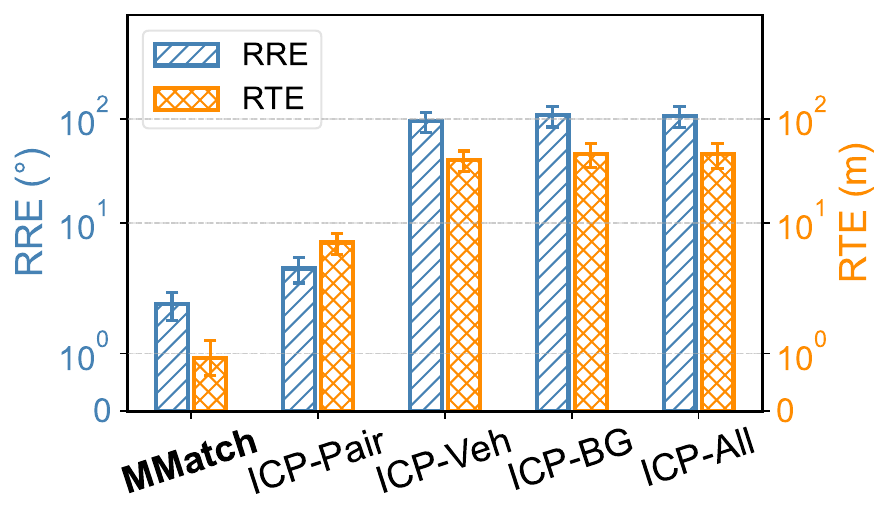}
		\caption{Performance of ablation experiments on real-world dataset.}
		\label{fig:real_abl}
	\end{minipage}
	\hfill
	\begin{minipage}[t]{0.24\linewidth}
		\centering
		\includegraphics[width=1\linewidth]{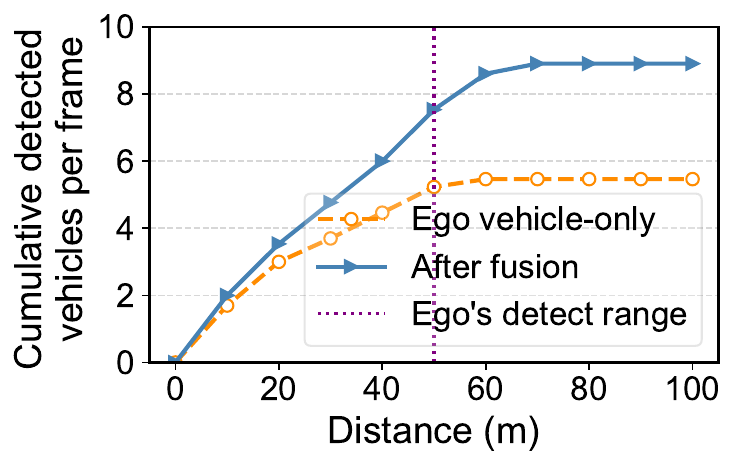}
		\caption{Performance of perception extension.}
		\label{fig:distance_extend}
	\end{minipage}
	\hfill
	\begin{minipage}[t]{0.24\linewidth}
		\centering
		\includegraphics[width=1\linewidth]{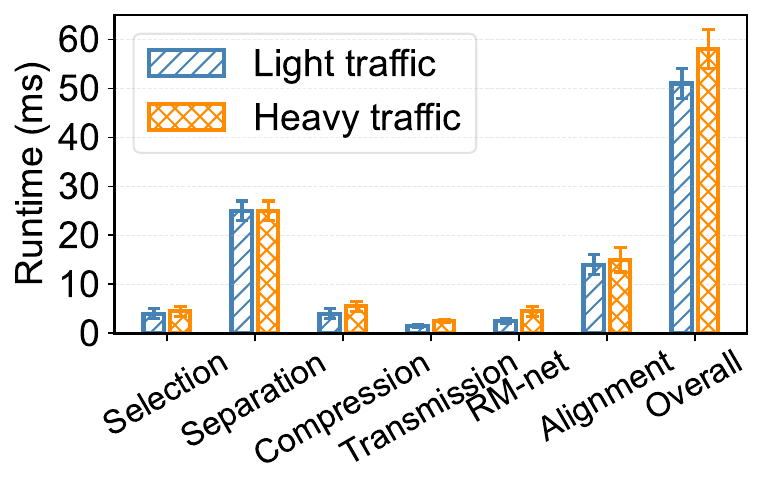}
		\caption{Performance of time latency.}
		\label{fig:time_latency}
	\end{minipage}
\end{figure*}
\textbf{Ablation experiments.} In addition, we also conducted ablation experiments to demonstrate the advantages of integrating co-visible PCD pairs and background radar. To this end, we compare the performance with the following four experiments, including ICP-All, ICP-BG, ICP-Veh, and ICP-Pair. In particular, in the ICP-All, we simply input all PCDs and perform ICP alignment. In ICP-BG and ICP-Veh scenarios, we only align the PCDs by the background and vehicles. As for ICP-Pair experiments, we first detect the co-visible vehicle PCD pairs by our learning model, and then align them with ICP. As presented in Fig.~\ref{fig:ablation_acc}, we find that the performance in ICP-Veh experiments is much worse than any other results in the easy mode. That is because the radar points covering the vehicles are much sparser than the background, and thus simply performing ICP leads to a large mismatch. In the hard mode, ICP-All and ICP-BG achieve worse results than other groups, which indicate that the symmetry of the background has a negative impact on the alignment. Compared with all the experiments, our background-constrained alignment algorithm makes both the PCD pairs and background PCDs contribute positive impacts on the view alignment.

\subsection{Alignment Accuracy on Real-world Dataset}\label{sec:re_accuracy} 
\textbf{Performance comparison.} In this section, we evaluate the performance on the real campus traffic. We first present an overall accuracy to show the alignment performance. As shown in Fig.~\ref{fig:real_bas}, MMatch achieves alignment accuracy in RRE about $1.8^\circ$ and RTE less than 0.9m, which is worse than the results in CARLA dataset. The reason is that mmWave signals suffer from multi-path reflection and the environmental noise in the real communication channel, even though we have filtered the outliers. However, MMatch still outperforms the recent alignment algorithm and achieves localization accuracy at the decimeter level, which enables many autonomous driving tasks. In addition, we also conducted the ablation experiments on the real-world dataset. According to the results presented in Fig.~\ref{fig:real_abl}, we can find that the proposed background-constrained alignment algorithm can still work in real traffic scenarios.

\textbf{Perception extension.} In this section, we conduct experiments to demonstrate the benefits of perception fusion in real-world traffic situations. Fig.~\ref{fig:distance_extend} presents the cumulative number of the detected vehicles within the distance of 100m from the Ego vehicle. According to the results, we can find that Ego vehicle can only detect about 6 vehicles on average. However, after perception fusion, cooperative CAV can help to increase the visible targets into about 9 vehicles. Assuming the common perception range for a vehicle is 50m, this extension results indicate that the perception range can be extended from 50m to 100m, which significantly improves the perception performance and helps avoid catastrophic accidents.

\subsection{Time Latency}\label{sec:time} 
In this section, we conduct the experiments on both the CARLA and treal-world datasets, where we employ the same data size from two datasets and calculate the average latency.

\textbf{Processing latency.}  As shown in Fig.~\ref{fig:time_latency}, under heavy traffic, our co-visible detection module based on \textit{RM-net} only requires 4.5ms. It is worth noting that the test processing of \textit{RM-net} rely on only the CPU. As for other modules, like radar selection, separation, image compression, data transmission, and view alignment, they create the time delay as 5.5ms, 25ms, 5.5ms, less than 3ms, and 15ms, respectively. In total, we have an overall time latency less than 59ms, which is faster than recent mainstream image-based tasks with a time latency of 100ms/frame. Therefore, MMatch can significantly satisfy the real-time requirement in many autonomous driving tasks.

\textbf{Communication overhead.} We also evaluate communication overhead by comparing the time consumption and data size with the approaches that share all the raw data and the image of vehicles. To emulate real channel states, we share image-radar frames in the campus traffic, where two 802.11ac WiFi routers are installed for transmission. As presented in Table~\ref{com_overhead}, MMatch requires sharing about 5-9KB and causes a time delay less than 3ms. In contrast, the raw data sharing approach needs 2765.8KB and 237ms in data size and time latency, while the image of objects without compression shares about 49.912-111.052KB and needs 10ms. Accordingly, MMatch reduces the data communication and time latency by about 395$\times$ and 118$\times$ compared to the raw data sharing system, and 12$\times$ and 5$\times$ compared to the image sharing. Therefore, MMatch is appropriate for deployment even in dynamic traffic conditions with low data rates.
\begin{table}[t]
	\centering
	\caption{Communication overhead.}
	\begin{tabular}{cccc}
		\toprule
		{} & \textbf{Data size (KB)}  &  \textbf{Overhead (ms)}   \\ 
		\midrule
		Raw data sharing & 2765.8  & 237    \\
		Image without compression  & 49.912-111.052 & 10  \\
		\textbf{MMatch}  & 5-9 & $<$3  \\ 
		\bottomrule
	\end{tabular}
	\label{com_overhead}\vspace{-0.3cm}
\end{table}

\subsection{Robustness}\label{sec:robustness} 
\begin{figure}[t]
	\hfill
	\begin{minipage}[t]{0.49\linewidth}
		\centering
		\includegraphics[width=1\linewidth]{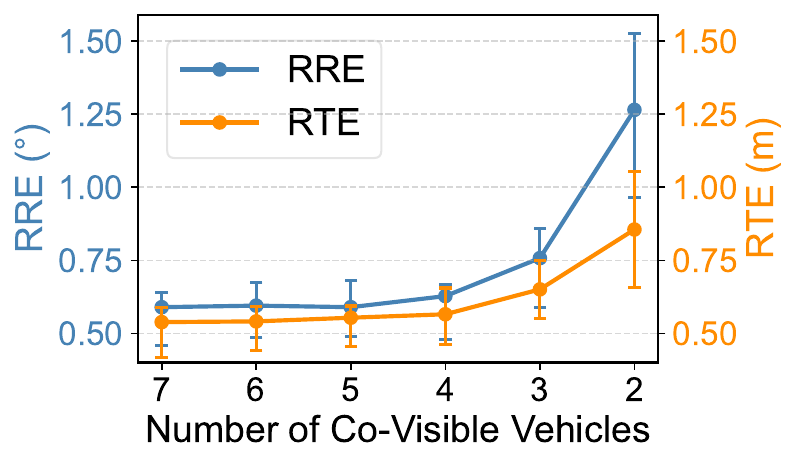}
		\caption{Impact of vehicle detection missing.}
		\label{fig:missing}
	\end{minipage}
	\hfill
	\begin{minipage}[t]{0.49\linewidth}
		\centering
		\includegraphics[width=1\linewidth]{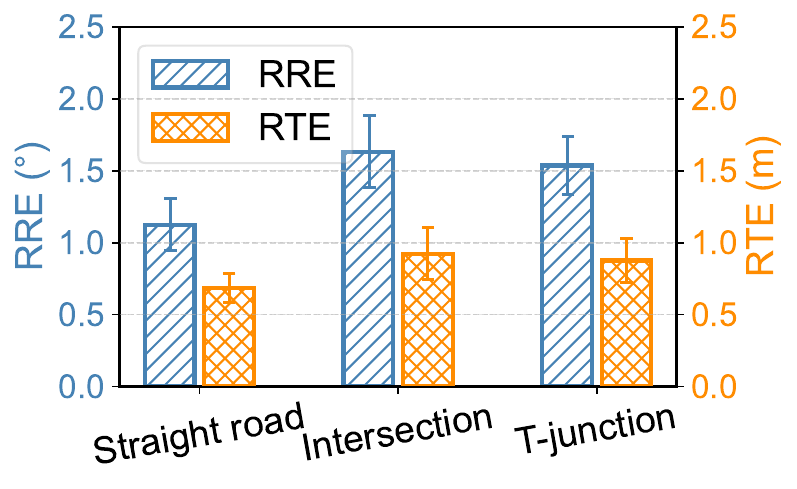}
		\caption{Performance on various road types.}
		\label{fig:road_type}
	\end{minipage}
\end{figure}

\textbf{Vehicle detection missing.} Due to the impact of the view angles or illumination changes, the Ego vehicle or cooperative CAV may have missed some targets in the view. As a result, the number of co-visible vehicles will decrease and thus lead to fewer PCD pairs for alignment. To this end, we show the impact of co-visible vehicle detection missing in the view. As shown in Fig.~\ref{fig:missing}, when we have 5 co-visible or more vehicles, MMatch can achieve the RRE less than $0.6^\circ$ and RTE less than 0.6m. When the number of missing vehicles increases, the transformation estimation errors also increase. However, we still achieve the RRE less than $1.4^\circ$ and RTE less than 1m, even though there are only two co-visible vehicles in the view. Therefore, MMatch requires only a small number of co-visible targets for alignment and it is robust to traffic changes.

\textbf{Road types.} In addition, we conduct our experiments in the most common straight road, intersection, and T-junction. In this experiment, we enable the Ego vehicle and CAV to move from the same and opposite directions in the straight road scenario. As for the intersection and T-junction experiments, we force the cooperative vehicles to move from various directions combinations. As presented in Fig.~\ref{fig:road_type}, MMatch achieves the RRE less than $2^\circ$ and RTE less than 1m for all these three scenarios.

\section{Related Work} \label{sec:related}

\subsection{Connected Vehicles and Infrastructures}\label{sec: connected} 
The development of vehicular communication makes cooperative perception among connected vehicles possible~\cite{aghashahi2021stochastic,moubayed2020edge,segata2022multi}. Recent advances attempt to directly exchange and fuse sensor data between CAVs through vehicle-to-vehicle (V2V) communications~\cite {qiu2018avr}. To achieve higher bandwidth and computation resources, other studies also explore vehicle-to-infrastructure (V2I) communication to improve perception fusion~\cite{he2022automatch,shi2022vips,he2021vi}. Compared with V2V, cooperative perception in V2I requires dense deployments of the infrastructures, which will be a huge investment~\cite{qiu2021autocast}. Besides, vehicles would fail cooperative perception in a strange environment without infrastructure deployments. In our design, MMatch can be deployed to both of V2V and V2I systems. 

\subsection{Perception Fusion}\label{sec: coo}
\textbf{Point-level perception fusion.} To enable perception fusion, PCD alignment have been widely explored to directly align the raw data by performing ICP or its variants~\cite{besl1992method,fang2022lidar,gao2019filterreg}. Thanks to the high-density of LiDAR, these approaches achieve high accuracy. However, the limited bandwidth may not support a large amount of raw data sharing in the context of cooperative perception, which poses significant challenges for the real-world autonomous driving system. To lower the latency, recent advantages design to separate the PCDs of targets and transmit only parts of the points for view alignment~\cite{chen2019cooper}. Nevertheless, these alignment methods require a large overlap and relative complete structures of objects, which is not appropriate for the vehicles that are located in large different views.

\textbf{Feature-level perception fusion.} Feature-level perceptions share only the representations, from which the communication overhead can be significantly reduced. Several studies have proposed to encode the raw sensor data and extract the spatial features~\cite{vadivelu2021learning} and planar structures~\cite{yuan2022leveraging} from the dense LiDAR points. However, they require homogeneous feature descriptors, which are sensitive to the environment. Other systems explore to extract landmark keypoints of the ground signs~\cite{he2022automatch}~\cite{liu2019fusioneye} from the image frames. Unfortunately, the textures, like the ground signs and the lane line are easily occluded by the vehicles, especially in a crowded traffic. Moreover, the image-based system lacks of the depth information.

\textbf{Object-level perception fusion.} These studies transmit only high-level results detected from the raw data, which is communication-friendly. Recent approaches attempt to treat the target as points and align them from different views~\cite{pomerleau2015review,billings2015generalized}. Nevertheless, they are applied for static targets and are computation resource-hungry, which is not suit for moving vehicles and cannot meet the real-time requirements. The approach in~\cite{shi2022vips} aligns the views by comparing the similarity of detected objects from the LiDAR. However, local similarity may lead to a mismatch of the co-visible vehicles. Besides, it relies in relative complete structures and accurate velocity estimation, which is unavailable for radar system. 

\section{Conclusion}\label{sec:conlude}
In this paper, we present MMatch, a lightweight system that can accurately and real-time fuse multi-vehicle perception and localize vehicles in its view with  mmWave radar assistance. MMatch constructs unique associations for all the targets with radar spatial information to find the co-visible vehicles for perception fusion without relying on raw data sharing. We implement MMatch on both datasets collected from the CARLA platform and real-world traffic. Experimental results demonstrate that MMatch achieves decimeter-level accuracy within 59ms in various traffic conditions, which significantly improve the safety for autonomous driving. 

\balance
\bibliographystyle{IEEEtran}
\bibliography{sample-bibliography}

\end{document}